\documentclass[sigconf,nonacm]{acmart}

\AtBeginDocument{%
  }

\setcopyright{acmlicensed}
\copyrightyear{2025}
\acmYear{2025}
\acmDOI{XXXXXXX.XXXXXXX}
\acmConference[]{}{}{}

\usepackage{multirow}
\usepackage{tabularx}
\usepackage{array}
\usepackage{listings}

\begin{document}

\title{English Pronunciation Evaluation without Complex Joint Training: LoRA Fine-tuned Speech Multimodal LLM}




\author{
{\bf Taekyung Ahn${^1}$${^,}$${^2}$, Hosung Nam${^2}$} \\
Enuma, Inc.${^1}$, Korea University${^2}$ \\
\texttt{taekyung@enuma.com}, \texttt{hnam@korea.ac.kr}
}

\renewcommand{\shortauthors}{Ahn et al.}

\begin{abstract}
This study demonstrates that a Multimodal Large Language Model (MLLM) adapted via Low-Rank Adaptation (LoRA) can perform both Automatic Pronunciation Assessment (APA) and Mispronunciation Detection and Diagnosis (MDD) simultaneously. Leveraging Microsoft's Phi-4-multimodal-instruct, our fine-tuning method eliminates the need for complex architectural changes or separate training procedures conventionally required for these distinct tasks. Fine-tuned on the Speechocean762 dataset, the pronunciation evaluation scores predicted by the model exhibited a strong Pearson Correlation Coefficient (PCC > 0.7) with human-assigned scores, while achieving low Word Error Rate (WER) and Phoneme Error Rate (PER) (both < 0.15). Notably, fine-tuning only the LoRA layers was sufficient to achieve performance levels comparable to those achieved by fine-tuning all audio layers. This research highlights that an integrated pronunciation assessment system can be established by adapting large multimodal models without full fine-tuning, utilizing a significantly simpler training methodology compared to previous joint models designed for simultaneous APA and MDD. This efficient LoRA-based approach paves the way for more accessible, integrated, and effective Computer-Assisted Pronunciation Training (CAPT) technologies for English L2 learners.
\end{abstract}




\maketitle

\begin{figure}[h]
  \centering
  \includegraphics[width=.9\linewidth]{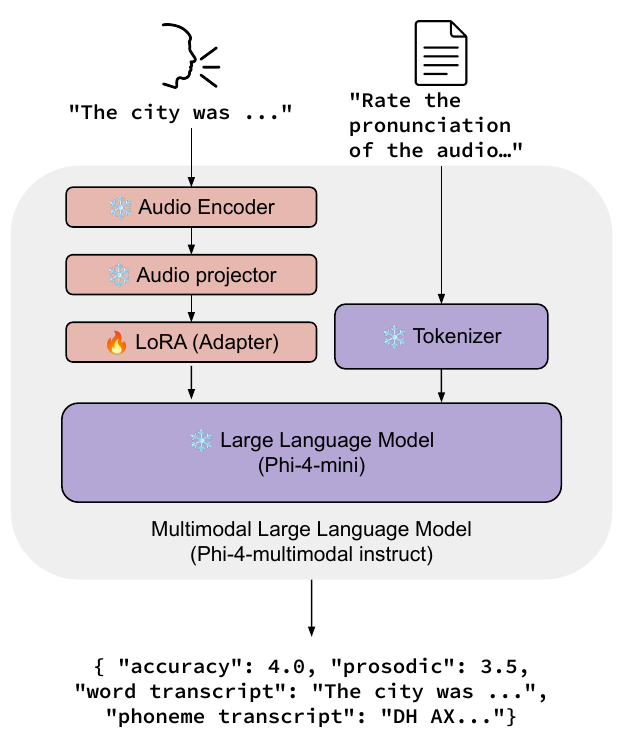}
  \caption{The overview of our proposed method.}
  \label{fig:freeze}
  \Description{}
\end{figure}

\section{Introduction}
With the advancement of ASR (Automatic Speech Recognition) technology, various technologies have been developed in the CAPT (Computer-assisted pronunciation training) field to help non-native English speakers with English pronunciation \cite{fouz2015trends}. CAPT is a system that automatically evaluates pronunciation in speech. It can be primarily divided into the APA (Automatic Pronunciation Assessment) task, which evaluates given utterances with scores, and the MDD (Mispronunciation Detection and Diagnosis) task, which detects whether there were mispronunciations when a speaker reads a given sentence.

The emergence of end-to-end pre-trained speech recognition models has enabled superior performance through fine-tuning with minimal data and streamlined training processes \cite{baevski2020wav2vec} \cite{babu2022xls} \cite{conneau2021unsupervised}. CAPT systems leveraging these models have advanced in two primary directions: enhancing mispronounced speech recognition through fine-tuning techniques, and developing robust pronunciation evaluation frameworks based on recognition outcomes \cite{peng2021study} \cite{yang2022improving} \cite{getman2022wav2vec2}. The inherent flexibility of these models has facilitated research on unified systems that simultaneously perform APA and MDD tasks \cite{ryu2023joint}, revealing a significant correlation between these two functionalities. However, these approaches still require separate datasets and independent model architectures for each task, necessitating distinct training procedures and computational resources despite their functional interdependence.

Recently, LLMs (Large Language Models) have exhibited significant advancements in understanding the context of natural language \cite{heo2024llm}. LLMs can perform the required purpose through a small amount of additional training or even aligning using only prompts, without gathering massive training data and computing resources. Following these advancements, multimodal LLMs (MLLMs) combined with vision and audio encoders have emerged. They employ an architecture that connects encoders to existing LLMs, enabling the conversion of raw images or audio files into vector representations similar to text embeddings. By training adapters to understand this vector, LLMs can understand image or audio inputs in the same context as text.

To address the limitations of existing separate training approaches, our research introduces a unified method that executes both APA and MDD tasks within a single training framework and model architecture. We specifically employ Phi-4-multimodal-instruct \cite{abouelenin2025phi}, leveraging its MLLM capabilities to overcome the traditional requirement for independent datasets and model layers. 
By utilizing efficient LoRA (Low-Rank Adaptation) fine-tuning that learns only a speech adapter, our approach bypasses the computational demands of full fine-tuning while maintaining performance. Our objective is to demonstrate that this unified MLLM framework can effectively handle both pronunciation evaluation tasks simultaneously, providing a more resource-efficient and streamlined alternative to existing multi-stage CAPT systems.

\section{Related Works}
\subsection{CAPT system with ASR}
CAPT (Computer-assisted pronunciation training), emerged in the 1980s \cite{fouz2015trends}. In CAPT, pronunciation education has been actively used and researched by incorporating ASR (automatic speech recognition) technology. Pronunciation evaluation methodology has been researched since the 1990s \cite{hong2021automated} \cite{bernstein1990automatic} \cite{franco2000combination}, and this area can be largely divided into Automatic Pronunciation Assessment (APA) and Mispronunciation Detection and Diagnosis (MDD) tasks.

APA refers to a series of methodologies that help language learning by building a system that automatically scores learners' utterances. With ASR technology, it was possible to build a system that could recognize phoneme units of speech and automatically score pronunciation accuracy by comparing them with the correct phonemes \cite{franco2000sri} \cite{franco2000combination}. This research has been conducted in various ways regarding evaluation methods that can measure pronunciation accuracy with a metric called goodness of pronunciation (GOP) using HMM-based ASR systems \cite{hong2021automated} \cite{witt2000phone}.

MDD research proceeded by comparing extracted phoneme-level acoustic features with correct phonemes. Traditional mispronunciation detection and diagnosis technologies typically rely on expert-crafted phonological rules, statistical methods such as Hidden Markov Models (HMMs) and Gaussian Mixture Models (GMMs), and acoustic feature analysis including MFCCs and formant tracking, which require extensive manual engineering and often struggle with context-dependent variations, prosodic features, and speaker variability  \cite{witt2000phone}  \cite{eskenazi2009overview}  \cite{qian2010use}.

\subsection{End-to-end ASR modeling}

\begin{figure}[h]
  \centering
  \includegraphics[width=.9\linewidth]{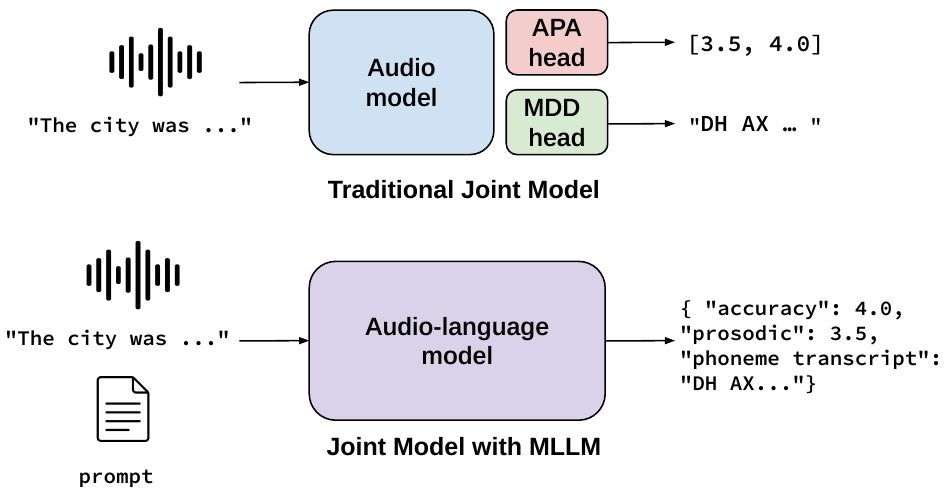}
  \caption{The Comparison of our model and past studies.}
  \label{fig:overview}
  \Description{}
\end{figure}

\begin{figure*}[h]
  \centering
  \includegraphics[width=\textwidth]{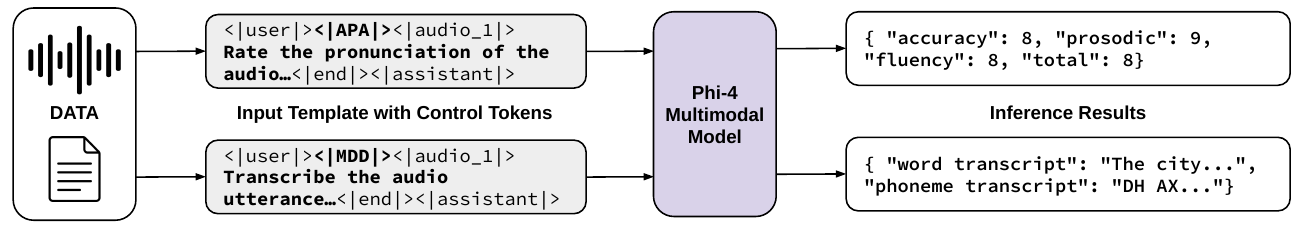}
  \caption{The methods for using the prompts and control tokens.}
  \label{fig:controltoken}
\end{figure*}

With advances in machine learning technology, speech recognition models can now be trained without complex preprocessing procedures as deep learning models can autonomously learn the relationship between speech and text \cite{feng2020sed}. Gong \cite{gong2022transformer} proposed GOPT, a Transformer model based on GOP features for assessing non-native English speakers' pronunciation. This model simultaneously evaluates multiple aspects of pronunciation quality (accuracy, fluency, completeness, prosody) across different features (phoneme, word, utterance) through multi-task learning, improving performance for each assessment task. Experiments on the Speechocean762 \cite{zhang2021speechocean762} dataset demonstrated that GOPT significantly outperforms previous traditional methods in pronunciation assessment.

With the emergence of SSL (self-supervised learning) methods such as wav2vec2.0 \cite{baevski2020wav2vec}, ASR models can learn speech features as units smaller than phonemes. Researchers could easily build APA and MDD engines by fine-tuning with a small amount of data \cite{peng2021study} \cite{yang2022improving} \cite{getman2022wav2vec2}. Despite these advances, most approaches still require separate training procedures for APA and MDD tasks. \cite{ryu2023joint} built a joint model that could perform both APA and MDD together by fine-tuning the wav2vec2.0 model with speech data transcribed at the phoneme level, then performing multi-task learning with separated APA and MDD heads. However, this approach still necessitated separate datasets and distinct training processes for each task encoder, limiting scalability and resource efficiency. Additionally, being trained solely on speech input, these models cannot incorporate other modalities such as text prompts for enhanced flexibility.

\subsection{Multimodal Large Language Model (MLLM)}
Subsequently, Multimodal Large Language Models (MLLMs) emerged with the ability to process images and speech alongside text. These models convert visual or audio inputs into vector dimensions through specialized encoders, then align these vector representations with LLM embeddings to enable comprehensive multimodal understanding.
Wang \cite{wang2025exploring} investigated the zero-shot capabilities of the GPT-4o \cite{hurst2024gpt} for pronunciation assessment, evaluating its performance on multi-level (phoneme, word, sentence) scoring. The study shows GPT-4o's zero-shot scoring accuracy was found to be significantly lower than other evaluation methods. This study concludes that MLLMs like GPT-4o may not yet replace specialized tools for scoring pronunciation without fine-tuning.

Fu \cite{fu2024pronunciation} developed a pronunciation evaluation system based on MLLMs. In their research, they trained a dedicated audio encoder and integrated it with an LLM, achieving high Pearson Correlation Coefficient (PCC) values between the system's predicted pronunciation scores and human evaluators. This study successfully demonstrated the viability of MLLM-based pronunciation assessment systems. However, this study used approximately 1,000 hours of speech data to train the audio encoder for integration with the LLM. Connecting pre-trained audio encoders and LLMs through adapters to create MLLMs presented significant challenges. This approach involved complex integration processes resulting in substantial costs for training and deployment. 

\subsection{Phi-4-multimodal with LoRA fine-tuning}
Phi-4-multimodal \cite{abouelenin2025phi} is a visual-speech language model released by Microsoft, a multimodal model that connects visual and speech encoders to LLM. Phi-4 incorporates a Mixture-of-LoRAs (Low-Rank Adaptations) structure in the network, designing the architecture to achieve good performance while minimizing interference between modalities by training only LoRA layers, not full fine-tuning. With this advantage, various fine-tuning can be conducted with limited computational resources compared to the model size.

Leveraging these capabilities of Phi-4-multimodal, our research addresses the fundamental limitations of existing CAPT systems that require separate training procedures and datasets for APA and MDD tasks. By utilizing the unified MLLM framework with efficient LoRA fine-tuning, we demonstrate that both tasks can be accomplished simultaneously within a single model architecture and training process. This approach not only eliminates the resource-intensive requirements of previous methods but also enables flexible system customization through prompt engineering, representing a significant advancement toward more accessible and efficient CAPT technologies. Our work establishes the feasibility of building a truly unified pronunciation assessment system that overcomes the computational and architectural constraints that have historically separated these functionally related tasks.

\section{Method}
\subsection{Control Tokens}
 To differentiate between tasks without modifying the model architecture, we employed control tokens: $<$$|$APA$|$$>$ for APA prompts and $<$$|$MDD$|$$>$ for MDD prompts (see Figure \ref{fig:controltoken}). These tokens were prepended to each prompt before Supervised Fine-Tuning (SFT). During the SFT process, the model learned to associate each control token with the specific requirements of the corresponding task. We then experimentally evaluated the performance benefits during inference when explicitly guiding the model using these task-specific control tokens. For baseline comparison, we used the Phi-4-multimodal-instruct model \cite{abouelenin2025phi} for inference without control tokens or additional training.

\begin{figure}[h]
  \centering
  \includegraphics[width=\linewidth]{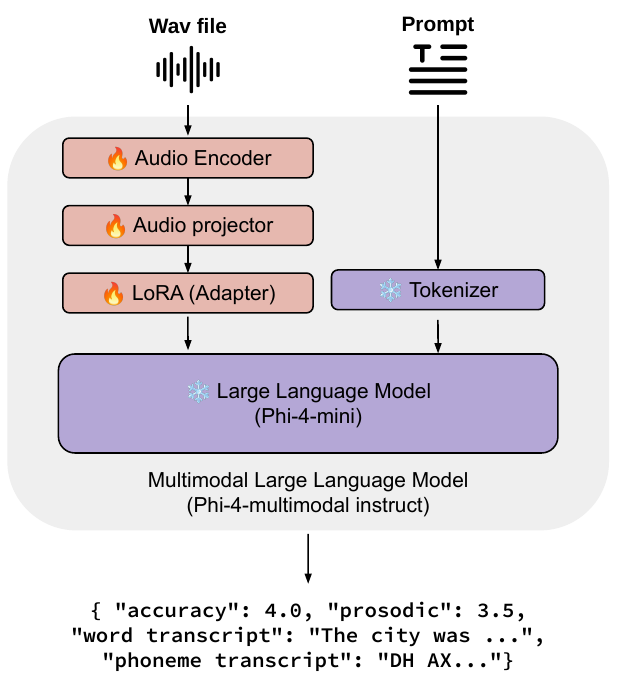}
  \caption{The method of unfreezing layers.}
  \label{fig:unfreeze}
  \Description{}
\end{figure}

\subsection{Encoder Layer Unfreeze}
This study investigated the effect of unfreezing specific audio-related layers. Using the same dataset, we compared two distinct fine-tuning strategies:

1. LoRA-only fine-tuning: Updating only the weights of the LoRA adapter layers, while keeping other layers, including the Audio Encoder and Audio Projector, frozen (See Figure \ref{fig:freeze}).

2. Unfreeze fine-tuning: Updating the weights of the LoRA adapter layers in addition to unfreezing and updating the weights of the Audio Encoder and Audio Projector layers (See Figure \ref{fig:unfreeze}).

While conventional practice typically limits fine-tuning to LoRA adapter layers, the official documentation \cite{microsoft2025phi} suggests potential benefits from unfreezing all audio layers. For this reason, our experiment directly compares these approaches to determine whether unfreezing the Audio Encoder and Projector alongside the LoRA adapter provides performance advantages for audio-related tasks.

\subsection{Find the Correlation between APA and MDD tasks}
Based on Ryu et al.'s study \cite{ryu2023joint}, we analyzed the correlation between pronunciation accuracy scores and  pronunciation recognition performance. We examined whether scores from the single model (jointly trained on both APA and MDD tasks using identical utterance datasets) genuinely reflect pronunciation quality rather than merely projecting data to scores. Specifically, we investigated the relationship between human and model's accuracy scores and Phoneme Error Rate (PER) from the MDD task. We hypothesized a negative correlation, where higher accuracy scores would correspond to lower PER, suggesting that pronunciation assessment is indeed informed by mispronunciation detection.

\section{Experiments}
\subsection{Dataset}

\begin{table}[t]
\centering
\caption{Speechocean762 Dataset Information}
\label{tab:speechocean762_data}
\begin{tabular}{llcc}
\toprule
\multicolumn{2}{l}{\textbf{Type (Size)}} & \textbf{Train (2.5k)} & \textbf{Test (2.5k)} \\
\midrule
\multicolumn{2}{l}{Total files} & 2500 & 2500 \\
\multicolumn{2}{l}{Speakers} & 125 & 125 \\
\multicolumn{2}{l}{Total wav duration} & 2h 52min & 2h 41min \\
\midrule
\multirow{5}{*}{\textbf{Age (n)}} & under 20 & 49.6\% (62) & 56.0\% (70) \\
& 20s & 44.0\% (55) & 36.0\% (45) \\
& 30s & 6.4\% (8) & 7.2\% (9) \\
& 40s & - & 0.8\% (1) \\
& over 50 & - & - \\
\midrule
\multirow{2}{*}{\textbf{Gender(n)}} & male & 46.4\% (67) & 53.6\% (58) \\
& female & 53.6\% (58) & 46.4\% (67) \\
\bottomrule
\end{tabular}
\end{table}

To validate the model's performance constructed with our proposed learning approach and enable comparison with prior research, we conducted experiments utilizing the Speechocean762 dataset \cite{zhang2021speechocean762}. We used the entire data, comprising 2.5k samples for the training set and 2.5k samples for the test set (see Table \ref{tab:speechocean762_data}).

The Speechocean762 dataset consists of English speech recordings produced by non-native speakers whose first language is Mandarin Chinese. For APA tasks, the dataset provides five sentence-level evaluation metrics: accuracy, fluency, prosody, completeness, and total score, with each metric scored on a scale from 0 to 10. We excluded the 'completeness' metric from our training and evaluation as all test set values were uniformly 10, preventing meaningful correlation analysis with model predictions.

Furthermore, Speechocean762 provides supplementary transcriptions that pinpoint mispronounced phones, available for MDD model training. These annotations utilize a 46-unit phone set derived from the CMUDict standard \cite{weide1998carnegie}. The phone set comprises 39 phones in ARPABET format, an <unk> tag for unknown phones, and six additional phones specifically designed to capture second-language (L2) pronunciation patterns.

\subsection{Model Configuration}
Phi-4-multimodal-instruct model \cite{abouelenin2025phi} is structured with an audio encoder, projector, and LoRA adapter connected to a Large Language Model (LLM). According to the study, the audio encoder and projector have been sufficiently trained on significant multilingual speech data, eliminating the need for additional training \cite{abouelenin2025phi}. The study argues that only the LoRA adapter requires fine-tuning to perform specific tasks. Therefore, in this experiment, we plan to build a unified APA and MDD system by conducting LoRA fine-tuning of the  model. We used one NVIDIA A100 SXM (80GB VRAM) GPU for fine-tuning. In batch size 8, we set the gradient accumulation step to 8, using the Adam optimiser with an initial learning rate of 2 × 10-5.

\begin{table}[t]
\centering
\small
\sloppy
\caption{Training Values}
\label{tab:training_values}
\begin{tabularx}{\linewidth}{p{0.8cm} p{1cm} l >{\raggedright\arraybackslash}X}
\toprule
\textbf{Type} & \textbf{Key} & \textbf{Data Type} & \textbf{Value} \\
\midrule
\multirow{4}{*}{\begin{tabular}[t]{@{}l@{}}APA\end{tabular}}
& Accuracy & Integer & Accuracy of pronunciation. Accuracy of consonants and vowels. \\
\cmidrule(lr){2-4}
& Prosodic & Integer & Fluency of prosody. Intonation, stress, rhythm. Speech rate. Pauses in speech. \\
\cmidrule(lr){2-4}
& \begin{tabular}[t]{@{}l@{}}Fluency \\ \end{tabular} & Integer & How naturally and smoothly a speaker communicates without noticeable pauses, repetition, or stammering in their speech. \\
\cmidrule(lr){2-4}
& \begin{tabular}[t]{@{}l@{}}Total \\ \end{tabular} & Integer & An overall assessment of the pronunciation quality, considering all aspects of the speech. \\
\midrule
\multirow{2}{*}{\begin{tabular}[t]{@{}l@{}}MDD \\ \end{tabular}}
& Phoneme transcript & String & Transcribed phoneme-level text where the model recognized the input speech (CMUDict corpus). \\
\cmidrule(lr){2-4}
& Word transcript & String & Transcribed word-level text where the model recognized the input speech. \\
\bottomrule
\end{tabularx}
\end{table}

As shown in Table \ref{tab:training_values}, the model was configured to predict multiple evaluation metrics. For the APA task, it was trained to measure phoneme accuracy, prosodic quality, fluency, and total score for each utterance. Additionally, in the MDD task, the model performed both phonemic and orthographic transcription of the same utterances. While traditional MDD approaches typically focus solely on phoneme-level recognition to identify mispronunciations, our approach differed. Since our pre-trained model already possessed robust ASR capabilities, we designed it to simultaneously recognize both the correct sentence (orthographic transcription) and the actual phonemes as they were heard (phonemic transcription). 

Based on the configuration outlined in Table \ref{tab:training_values}, we constructed specific prompts for supervised fine-tuning (SFT) tailored to both APA and MDD tasks (see Appendix \ref{app:apa-speechocean} and \ref{app:mdd}). Following the Phi-4 training script format, the prompts, including control tokens and audio tokens, constitutes the 'user' component of each training pair (see Figure \ref{fig:controltoken}), while the training values (Table \ref{tab:training_values}) are converted into 'assistant' component using the identical JSON format prescribed in the prompts. These user-assistant pairs, combined with their corresponding vectorized audio files, create complete training instances for the SFT process.

\begin{table*}[t]
\centering
\caption{Training Results. All p-values for APA tasks are under 0.05, except for underlined values}
\label{tab:speechocean_results_pcc_mdd_combined} 
\bigskip
\begin{tabular}{@{}c@{}c@{\hspace{6pt}}c@{\hspace{6pt}}c@{\hspace{6pt}}c@{\hspace{6pt}}c@{\hspace{6pt}}c@{\hspace{6pt}}c@{\hspace{6pt}}c@{\hspace{6pt}}c@{\hspace{6pt}}c@{}}
\toprule
& & \multicolumn{4}{c}{\textbf{APA task - PCC values}} & \multicolumn{5}{c}{\textbf{MDD task Metrics}} \\
\cmidrule(lr){3-6} \cmidrule(lr){7-11}
\begin{tabular}[c]{@{}c@{}}\textbf{Training} \\ \textbf{speech layers}\end{tabular} &
\textbf{epoch} &
\textbf{Accuracy} & \textbf{Fluency} & \textbf{Prosodic} & \textbf{Total} &
\textbf{WER} & \textbf{PER} & \textbf{F1-score} & \textbf{Precision} & \textbf{Recall} \\
\midrule
no training & 0 &
-0.041 & \underline{-0.017} & -0.0493 & -0.104 &
0.97 & 0.792 & 0.143 & 0.154 & 0.134 \\
\addlinespace
LoRA & 1 &
0.547 & 0.585 & 0.567 & 0.544 &
0.15 & 0.137 & 0.686 & 0.689 & 0.682 \\
\addlinespace
LoRA & 2 &
0.637 & 0.726 & 0.709 & 0.662 &
\textbf{0.139} & 0.118 & 0.722 & 0.725 & 0.719 \\
\addlinespace
LoRA & 3 &
0.656 & 0.727 & 0.711 & \textbf{0.675} &
0.14 & \textbf{0.114} & \textbf{0.724} & \textbf{0.728} & \textbf{0.721} \\
\addlinespace
LoRA & 4 &
0.645 & \textbf{0.733} & \textbf{0.714} & 0.668 &
0.148 & 0.121 & 0.721 & 0.723 & 0.72 \\
\addlinespace
Unfreeze & 1 &
0.535 & 0.559 & 0.564 & 0.55 &
0.159 & 0.199 & 0.575 & 0.579 & 0.57 \\
\addlinespace
Unfreeze & 2 &
0.624 & 0.668 & 0.653 & 0.634 &
0.145 & 0.154 & 0.651 & 0.651 & 0.651 \\
\addlinespace
Unfreeze & 3 &
0.621 & 0.669 & 0.655 & 0.637 & 
0.148 & 0.15 & 0.663 & 0.665 & 0.662 \\
\addlinespace
Unfreeze & 4 &
\textbf{0.743} & 0.717 & 0.704 & 0.666 & 
0.142 & 0.142 & 0.667 & 0.671 & 0.663 \\
\bottomrule
\end{tabular}
\end{table*}

\subsection{Evaluation}
To evaluate the model's APA performance, we used Pearson Correlation Coefficient (PCC) to measure the correlation between human and model-predicted pronunciation scores. For MDD  evaluation, we assessed the model's speech recognition performance using Word Error Rate (WER). Additionally, for phoneme recognition results, we calculated Phoneme Error Rate (PER) and F1-score by comparing the model outputs with reference phoneme transcriptions to measure the accuracy of mispronunciation detection.
\begin{equation} \label{eq:wer}
\text{WER} = \frac{I + D + S}{N}
\end{equation}
\begin{equation} \label{eq:per}
\text{PER} = \frac{I + D + S}{N}
\end{equation}
The WER metric quantifies the performance of speech recognition models. In the WER formula (\ref{eq:wer}), \textit{I} represents the number of insertion errors, \textit{D} represents the number of deletion errors, \textit{S} represents the number of substitution errors, and \textit{N} represents the total number of words in the reference. The sum of these errors is divided by the total number of words to obtain the error rate. Similarly, PER (\ref{eq:per}) was calculated using the same approach as WER, but at the phoneme level rather than the word level.
\begin{equation}\label{eq:precision}
\text{Precision} = \frac{TP}{TP + FP}
\end{equation}
\begin{equation} \label{eq:recall}
\text{Recall} = \frac{TP}{TP + FN}
\end{equation}
\begin{equation} \label{eq:f1score}
\text{F1 score} = 2 \times \frac{\text{precision} \times \text{recall}}{\text{precision} + \text{recall}}
\end{equation}
In the precision and recall formulas (\ref{eq:precision} and \ref{eq:recall}), \textit{TP} represents True Positive (correctly identified mispronunciations), \textit{FP} represents False Positive (incorrectly flagged correct pronunciations), and \textit{FN} represents False Negative (missed mispronunciations). These metrics were then used to compute the F1-score (\ref{eq:f1score}), which provides a balanced measure of the model's mispronunciation detection capability.

\subsection{Results}
Table \ref{tab:speechocean_results_pcc_mdd_combined} summarizes the APA and MDD task training results. Both fine-tuning strategies significantly improved over the \textit{no training} baseline, which showed high initial error rates (WER 0.97, PER 0.792) and low detection performance (F1-score 0.143). The LoRA approach consistently yielded superior results compared to the \textit{Unfreeze} strategy across all metrics. The \textit{LoRA} approach achieved its lowest Word Error Rate (WER) of 0.139 at epoch 2 and its best Phoneme Error Rate (PER) of 0.114, F1-score (0.724), precision (0.728), and recall (0.721) at epoch 3. While the \textit{Unfreeze} strategy showed performance gains up to epoch 4, its best scores (WER 0.142, PER 0.142, F1-score 0.667) remained less optimal than those achieved with the \textit{LoRA} approach. These findings indicate that LoRA fine-tuning, particularly around epochs 2 and 3, was the more effective strategy for the MDD task rather than full audio layer fine-tuning in this experiment.

\begin{table}[t]
\centering
\caption{Comparison of the PCC Results with other studies.}
\label{tab:comparison_studies_star}
\bigskip
\begin{tabular}{lccccc} 
\toprule
\textbf{Model} & \textbf{accuracy} & \textbf{fluency} & \textbf{prosodic} & \textbf{total} & \textbf{PER} \\
\midrule
\begin{tabular}[t]{@{}l@{}}LoRA \\ (four epochs)\end{tabular} & 0.645 & 0.733 & 0.714 & 0.668 & 0.121 \\
\addlinespace
\begin{tabular}[t]{@{}l@{}}Unfreeze \\ (four epochs)\end{tabular} & \textbf{0.743} & 0.717 & 0.704 & 0.666 & 0.142 \\
\addlinespace
\begin{tabular}[t]{@{}l@{}}GOPT \\\cite{gong2022transformer} \end{tabular} & 0.714 & 0.753 & 0.760 & 0.742 & - \\
\addlinespace
\begin{tabular}[t]{@{}l@{}}Data2vec2 \\ \cite{fu2024pronunciation} \end{tabular} & 0.713 & \textbf{0.777} & - & - & - \\
\addlinespace
\begin{tabular}[t]{@{}l@{}}Joint-CAPT-L1 \\ \cite{ryu2023joint} \end{tabular} & 0.719 & 0.775 & 0.773 & 0.743 & \textbf{0.099} \\
\addlinespace
\begin{tabular}[t]{@{}l@{}}Azure PA \\ \cite{wang2025exploring} \end{tabular} & 0.7 & 0.715 & \textbf{0.842} & \textbf{0.782} & - \\
\bottomrule
\end{tabular}
\end{table}

\begin{figure*}[h]
  \centering
  \includegraphics[width=0.8\textwidth]{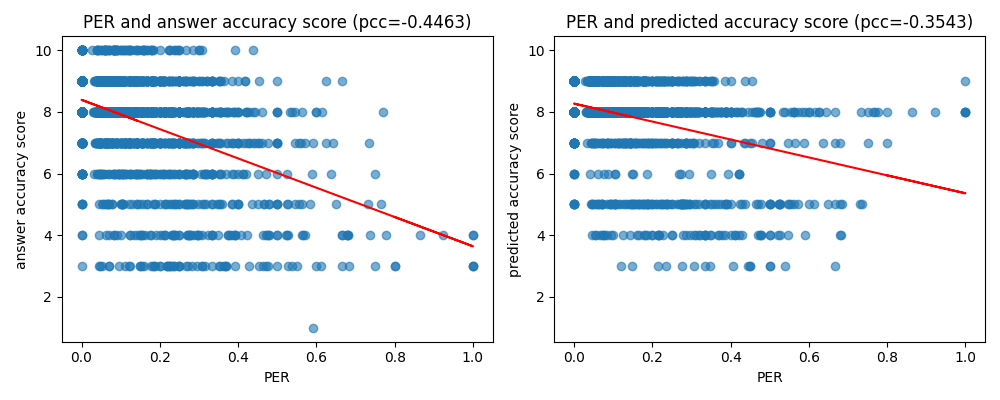}
  \caption{Pearson Correlation Coefficients between Accuracy Score and Phoneme Error Rate.}
  \label{fig:correlation_per_accuracy}
\end{figure*}

Table \ref{tab:comparison_studies_star} compares our fine-tuned models' performance (specifically \textit{LoRA} and \textit{Unfreeze} strategies evaluated at 4 epochs) on the Speechocean762 dataset against several established methods referenced in the literature. A notable achievement is that our \textit{Unfreeze} model yielded a state-of-the-art Pearson Correlation Coefficient (PCC) for accuracy (0.743), surpassing all other listed benchmarks including GOPT \cite{gong2022transformer}, Data2vec2 \cite{fu2024pronunciation}, Joint-CAPT-L1 \cite{ryu2023joint}, and Azure PA \cite{wang2025exploring} in this specific dimension.

However, this strength in accuracy assessment did not uniformly extend to other evaluated aspects of pronunciation. For fluency, both Data2vec2 and Joint-CAPT-L1 reported higher PCC values (0.777 and 0.775, respectively) than achieved by our models (\textit{LoRA}: 0.733, \textit{Unfreeze}: 0.717). Similarly, regarding prosody and the overall (total) assessment score, the Azure PA service demonstrated superior correlation (Prosodic PCC 0.842, Total PCC 0.782). The Total PCC scores from GOPT (0.742) and Joint-CAPT-L1 (0.743) also exceeded those of our \textit{LoRA} (0.668) and \textit{Unfreeze} (0.666) models. Furthermore, concerning the MDD aspect, measured here by PER, \cite{ryu2023joint} study achieved a lower (better) PER of 0.099. This indicates a higher phoneme-level recognition accuracy than our \textit{LoRA} model (0.121) and \textit{Unfreeze} model (0.142).

In summary, as detailed in Table \ref{tab:comparison_studies_star}, our proposed fine-tuning approach—particularly the \textit{Unfreeze} strategy—achieves state-of-the-art performance for phoneme-level accuracy scoring (Accuracy PCC) on the Speechocean762 benchmark. However, when evaluating broader dimensions like fluency and prosody, existing specialized systems and commercial services still outperform our models. These established solutions demonstrate superior capabilities in comprehensive pronunciation assessment, resulting in higher overall correlation scores and lower phoneme recognition error rates.

Figure \ref{fig:correlation_per_accuracy} presents a correlation analysis between PER and pronunciation accuracy scores. The left panel illustrates the relationship between PER and the answer (human) accuracy score, demonstrating a moderate negative correlation (Pearson Correlation Coefficient, -0.4463). This indicates that lower PER values generally correspond to higher answer accuracy scores. A similar trend was observed for the predicted accuracy score, as shown in the right panel. Although the correlation is slightly weaker (-0.3543), it confirms a distinct negative relationship between PER and the predicted accuracy score. Therefore, both analyses consistently suggest improved phoneme recognition (lower PER) is associated with higher accuracy for the ground truth answers and the model's predicted accuracy assessments.

\section{Discussion}
\subsection{Contribution}
This research is significant in that it demonstrates the ability to construct an MLLM-based pronunciation evaluation system that can perform both APA and MDD by fine-tuning only some layers of an MLLM. A pronunciation evaluation system can be built with a small amount of training data by creating prompts for training and inference without complex procedures. By utilizing the characteristic that the model can achieve sufficient performance by training only some speech-related layers despite its large size, we were able to reduce unnecessary training costs. The fact that spelling and phoneme units do not need to be trained separately also reduces unnecessary training steps. Since ARPABET pronunciation symbols consist of alphabets, we could efficiently build a system that performs spelling and phoneme-level recognition by providing them as prompts without needing to train or add phoneme tokens.

While this experiment built a model using an English-speaking dataset composed only of Mandarin L1 speakers, it would be easy to build a pronunciation evaluation system for English learners with various native languages. Since Phi-4's LLM learned text data in 23 languages and its audio layer learned speech data in 8 languages, it would be possible to build a pronunciation assessment model suitable for target speakers with a smaller dataset and a simple training method compared to training English speech data of EFL learners from scratch.

\subsection{Limitation}
Despite these promising results, our approach has several important limitations. The primary limitation stems from computational constraints that prevented us from conducting full fine-tuning experiments. Full fine-tuning involves updating parameters across the entire model architecture, including both the speech components and the core LLM, potentially offering superior adaptation capacity compared to modifying only specific layers or employing LoRA. Although existing work (e.g., \cite{abouelenin2025phi}) and our results demonstrate that LoRA tuning can achieve adequate performance, a comprehensive assessment and comparison would require full fine-tuning experiments, which were infeasible given our limited computing resources.

Another significant limitation concerns the absence of suprasegmental information in our pronunciation evaluation approach. We were unable to establish metrics for the model to learn prosodic features such as stress and intonation from our dataset. Consequently, we could not validate the effectiveness of prosodic and fluency scoring methods, such as examining correlations between accuracy scores and phoneme error rates (PER). However, with a dataset specifically designed to capture suprasegmental information, it would be feasible for the model to provide explanations for prosodic and fluency score calculations.

\subsection{Further Research}

Several avenues for future research emerge from this work. First, the computational requirements of our model present practical challenges for deployment. With approximately 5.8 billion parameters, the model exhibits slow inference speeds and demands high-performance GPUs. For practical implementation in educational settings, where fast response times and cost-effective solutions are essential for commercialization, model optimization through quantization becomes crucial. While quantization scripts are available in the official documentation, comprehensive experiments comparing performance before and after quantization are necessary to assess the trade-offs between model efficiency and accuracy.

Second, comprehensive pronunciation evaluation requires assessing prosodic elements alongside phoneme accuracy. Our current approach evaluates prosodic characteristics using a single prosodic score, which provides limited insight into specific prosodic features such as stress, intonation, and rhythm. Future research should focus on extracting detailed prosodic information from training datasets, enabling the model to understand and evaluate both phonemic accuracy and prosodic elements simultaneously. This enhancement would allow the MLLM to provide more sophisticated speech analysis and generate detailed explanations of prosodic patterns.

Third, the multimodal capabilities of the Phi-4-multimodal-instruct model \cite{abouelenin2025phi} offer promising opportunities for integrated analysis. This model supports joint training of vision and speech encoders, enabling simultaneous processing of visual and auditory information. 
By fine-tuning this architecture to recognize pronunciation patterns in conjunction with lip movements, we could achieve higher-dimensional understanding and analysis of speech production. Such an approach would leverage both acoustic and visual cues within a unified framework, potentially improving pronunciation assessment accuracy and providing more comprehensive feedback to learners.

\section{Conclusion}
This research establishes a unified framework for Computer-Assisted Pronunciation Training (CAPT) by integrating Automatic Pronunciation Assessment (APA) and Mispronunciation Detection and Diagnosis (MDD) within a single Speech Multimodal Large Language Model (MLLM).
Through efficient Low-Rank Adaptation (LoRA) fine-tuning of the pre-trained Phi-4-multimodal-instruct model, we achieved pronunciation assessment performance that strongly correlates with human evaluations while maintaining high accuracy in speech and phoneme recognition tasks. This work contributes a practical, unified framework that lowers technical barriers for developing advanced pronunciation training technologies.

\section{Appendix}
\lstset{
    breaklines=true,
    breakatwhitespace=true, 
    basicstyle=\linespread{1.0}\small\ttfamily, 
    showstringspaces=false,
    columns=flexible,
    captionpos=b, 
    frame=single 
}
\subsection{APA prompt}\label{app:apa-speechocean}
\begin{lstlisting}
Rate the pronunciation of the audio.

**Accuracy**
Score range: 0 - 10
* 9-10: The overall pronunciation of the sentence is excellent, with accurate phonology and no obvious pronunciation mistakes
* 7-8: The overall pronunciation of the sentence is good, with a few pronunciation mistakes
* 5-6: The overall pronunciation of the sentence is understandable, with many pronunciation mistakes and accent, but it does not affect the understanding of basic meanings
* 3-4: Poor, clumsy and rigid pronunciation of the sentence as a whole, with serious pronunciation mistakes
* 0-2: Extremely poor pronunciation and only one or two words are recognizable

**Fluency**
Score range: 0 - 10
* 8-10: Fluent without noticeable pauses or stammering
* 6-7: Fluent in general, with a few pauses, repetition, and stammering
* 4-5: The speech is a little disfluent, with many pauses, repetition, and stammering
* 0-3: Intermittent, very disfluent speech, with lots of pauses, repetition, and stammering

**Prosodic**
Score range: 0 - 10
* 9-10: Correct intonation at a stable speaking speed, speak with cadence, and can speak like a native
* 7-8: Nearly correct intonation at a stable speaking speed, nearly smooth and coherent, but with little stammering and few pauses
* 5-6: Unstable speech speed, many stammering and pauses with a poor sense of rhythm
* 3-4: Unstable speech speed, speak too fast or too slow, without the sense of rhythm
* 0-2: Poor intonation and lots of stammering and pauses, unable to read a complete sentence

**Total**
Score range: 0 - 10
Provide an overall assessment of the pronunciation quality considering all aspects of the speech. This should reflect your holistic evaluation of the speaker's pronunciation abilities based on the entire audio sample.
* 9-10: Excellent overall pronunciation that sounds nearly native-like
* 7-8: Good pronunciation with minor issues that don't affect comprehension
* 5-6: Fair pronunciation with noticeable non-native features but generally understandable
* 3-4: Poor pronunciation that requires effort to understand
* 0-2: Very poor pronunciation that is largely incomprehensible

Provide the results in the following JSON format:
{'accuracy': ACCURACY_SCORE, 'fluency': FLUENCY_SCORE, 'prosodic': PROSODIC_SCORE, 'total': TOTAL_SCORE}

\end{lstlisting}

\subsection{MDD prompt} \label{app:mdd}
\begin{lstlisting}
Transcribe the audio utterance, providing both a word-level transcript and phoneme-level breakdown.

For the phoneme breakdown, use the CMU Pronouncing Dictionary format (e.g., AA, IH).
If a word or phoneme is unclear, mark it with '<unk>'.

Provide the results in the following JSON format:
{'word_transcript': 'That's an interesting observation.', 'phoneme_transcript': 'DH EH S AX N IH N T AX R EH S T IH NG AA B Z AX R V EY IH SH AX N'}
\end{lstlisting}



\newpage
\bibliographystyle{ACM-Reference-Format}
\bibliography{references}

\end{document}